\pdfoutput=1

\documentclass[11pt]{article}
\usepackage{authblk}
\usepackage{korean_insurance_project}

\usepackage{times}
\usepackage{latexsym}
\usepackage[T1]{fontenc}
\usepackage[utf8]{inputenc}
\usepackage{microtype}
\usepackage{inconsolata}
\usepackage{graphicx}

\title{A Korean Legal Judgment Prediction Dataset for Insurance Disputes}

\author[1]{Alice Saebom Kwak}
\author[1]{Cheonkam Jeong}
\author[2]{Ji Weon Lim}
\author[3]{Byeongcheol Min}
\affil[1]{The University of Arizona}
\affil[2]{Hana TI}
\affil[3]{Sungkyunkwan University}
\affil[ ]{\tt {\{alicekwak, cheonkamjeong\}@arizona.edu,}}
\affil[ ]{\tt {\{ljw1858, bcmin1018\}@gmail.com}}

\begin{document}
\maketitle
\begin{abstract}

This paper introduces a Korean legal judgment prediction (LJP) dataset for insurance disputes. Successful LJP models on insurance disputes can benefit insurance companies and their customers. It can save both sides’ time and money by allowing them to predict how the result would come out if they proceed to the dispute mediation process. As is often the case with low-resource languages, there is a limitation on the amount of data available for this specific task. To mitigate this issue, we investigate how one can achieve a good performance despite the limitation in data. In our experiment, we demonstrate that Sentence Transformer Fine-tuning (SetFit, \citealp{tunstall2022efficient}) is a good alternative to standard fine-tuning when training data are limited. The models fine-tuned with the SetFit approach on our data show similar performance to the Korean LJP benchmark models \citep{hwang2022a} despite the much smaller data size.

\end{abstract}

\section{Introduction}

Legal judgment prediction (LJP) is a subfield of Artificial Intelligence (AI) and law that utilizes computational methods to predict judgment outcomes based on descriptions of factors, such as facts and precedents, automatically. LJP intersects with various areas of research fields; inter alia, Natural Language Processing (NLP) has been widely applied to it as LJP is mostly represented in textual form (e.g. legal cases, contracts). Although there are gaps in performance between humans and machine, a body of research has been conducted on LJP automation with NLP techniques to increase efficiency with less cost (e.g., \citet{chalkidis2019neural}; \citet{ma2021legal}, among others). It is especially true of fields where legal disputes frequently occur.

Data-driven approaches, such as deep learning models, require large-scale data. However, legal domain lacks benchmark datasets due to the nature of the domain: that privacy is of paramount importance. Thus, early works on LJP focused on building benchmark datasets. For instance, \citet{xiao2018cail2018} built a large-scale Chinese legal dataset, called the Chinese AI and Law challenge dataset (CAIL2018), with more than 2.6 million criminal cases from the Supreme People's Court of China. Similarly, \citet{chalkidis2019neural} established new English legal judgment prediction datasets, including cases from the European Court of Human Rights. Recently, a Korean legal dataset, LBOX OPEN, has been constructed with 147k Korean precedents \cite{hwang2022a}.

With the accessibility of large-scale datasets, though only a few, a rising body of work has been conducted on the LJP tasks, in particular one that predicts the court's judgment based on facts of legal cases manually written by judges (e.g., \citet{chalkidis2019neural}; \citet{chen2019charge}). These applications corroborate the adaptation of NLP to legal domain and the efficiency and practicality of the interface between LJP and NLP. However, lack of a variety of legal datasets for LJP has hindered its expansion to other tasks and domains.

Most of the previous LJP tasks were performed with precedents for predictions of trials due to the sources of the benchmark datasets. However, LJP is not limited to trials; rather, it can be expanded to any kind of legal conflicts. One case may be legal disputes, which are as much, maybe more, prevalent as trials as they are easier and faster to mediate than trials. Thus, the present study introduces a dataset for legal disputes, specifically Korean legal disputes in the field of insurance, considering its practicality. 

Our key contributions are as follows:

\begin{itemize}
    \item We present a new Korean legal judgment prediction (LJP) dataset for insurance disputes, which is a novel and highly practical task. The dataset consists of 473 insurance dispute cases (231K tokens).
    \item We confirm that SetFit is a good alternative to a standard fine-tuning in a low-resource setting. The model fine-tuned with the SetFit framework shows competitive performance compared to the Korean LJP benchmark model, despite the much smaller dataset size.
\end{itemize}

\begin{figure*}[hbt!]
  \includegraphics[width=\textwidth]{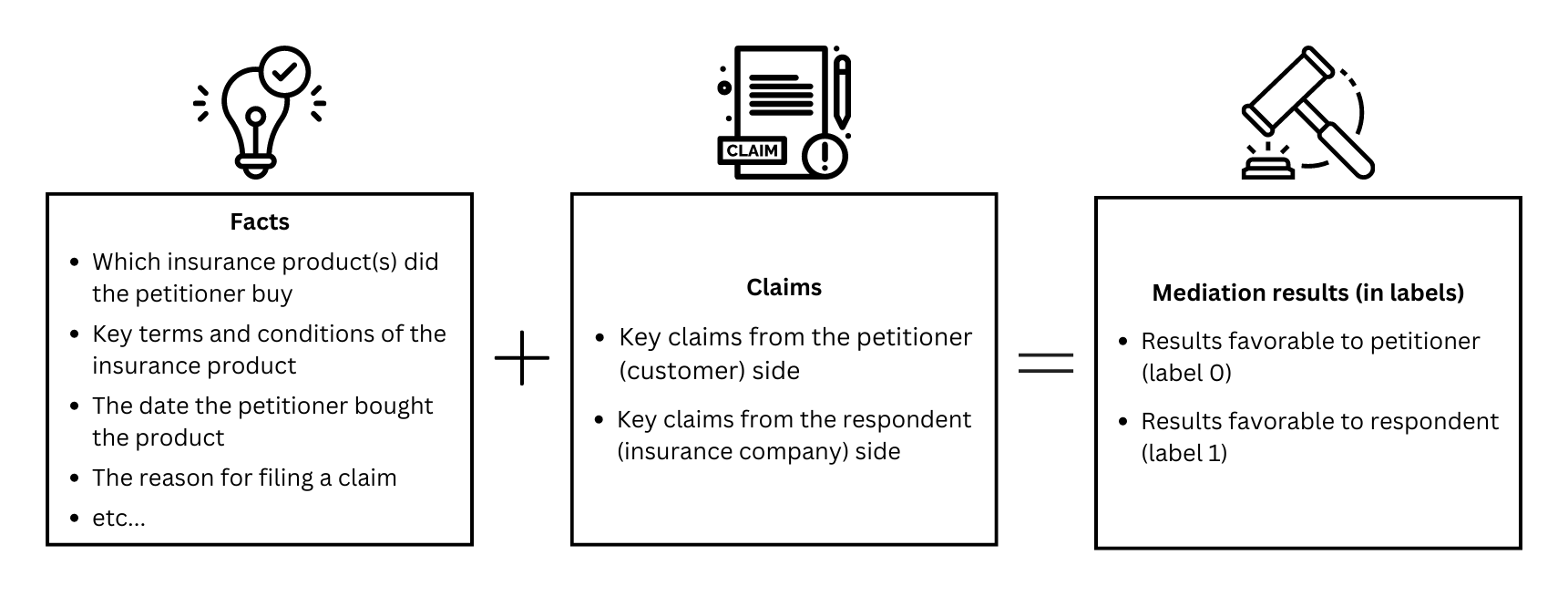}
  \vspace{-10mm}
  \caption{The figure above shows the structure of our dataset. The dataset consists of three parts: facts, claims, and mediation results. Facts and claims serve as the input texts and mediation results are the labels.}
\end{figure*}

\section{Dataset}

\begin{table}
\centering
\begin{tabular}{lll}
\hline
\textbf{Label} & \textbf{Cases} & \textbf{Tokens} \\
\hline
0 & 287 & 147437 \\
\hline
1 & 186 & 83633 \\
\hline
total & 473 & 231070 \\
\hline
\end{tabular}
\caption{\label{dataset statistics} The data statistics and the label distribution
}
\end{table}

We created a Korean legal judgment prediction dataset, consisting of 231K tokens from 473 insurance dispute cases provided by Financial Supervisory Service and Korea Consumer Agency. See Table 1 for the data statistics and the label distribution.

Our dataset consists of three parts: facts, claims, and mediation results. "Facts" part includes objective facts about the insurance dispute which both the petitioner and the respondent agree upon. "Claims" part contains key claims from each side. Lastly, the mediation results are labeled as either 0 or 1. If the mediation result is favorable to the petitioner, label 0 is given. On the other hand, if the result is favorable to the respondent, label 1 is given. All the other cases (e.g., cases with the results not favorable to either side) are excluded from the dataset. For further details about the dataset structure, see Figure 1.

\subsection{Data Collection}

We collect 493 insurance dispute cases, provided by Financial Supervisory Service and Korea Consumer Agency. These cases are publicly available on their official homepage. From the collected cases, we automatically extract facts and claims using regular expressions. Thereafter, we manually check whether the extracted sets of facts and claims match up. 

\subsection{Data Annotation}

After extracting the sets of facts and claims from all the cases, we annotate each case with binary labels (i.e., 0 or 1) based on its mediation result. If a mediation result is favorable to the petitioner side (i.e., customer), the case is labeled as 0. If a mediation result is favorable to the respondent side (i.e., insurance company), the case is labeled as 1. As the original data contain the official mediation results, we consider this information to be gold labels. While annotating, we exclude 20 cases, of which mediation results are not clearly favorable to either side. 

After the annotation is completed, we preprocess the data to remove special characters and Chinese characters. The original data contain various special characters, which are used to replace the personal information. We thus replace these special characters with the capital X. We also replace all the Chinese characters in the original data to the corresponding Korean characters by utilizing hanja library\footnote{https://github.com/suminb/hanja}. 

\begin{table*}[hbt!]
\centering
\begin{tabular}{lll}
\hline
\textbf{Models/Approaches} & \textbf{Accuracy} \\
\hline
paraphrase-mpnet-base-v2 + SetFit & 70.5 \\
Support Vector Machines & 64.5 \\
kobigbird-bert-base + Fine-tuning & 64.2 \\
Logistic Regression & 63.5 \\
Random Forest Classification & 62.5 \\
Naive Bayes & 58 \\
\hline
\textbf{LJP-Civil Lv1 Benchmark models} \citep{hwang2022a} \\
\hline
mt5-small + d.a. & 69.1 ± 0.1 \\
mt5-large (512 g) & 69.0 \\
LCUBE-medium & 68.9 \\
\hline

\end{tabular}
\caption{\label{evaluation results}
The evaluation results for the models trained/fine-tuned on our dataset
}
\end{table*}

\section{Evaluation}

The amount of the data available for our task is limited (473 data points), as is often the case for low-resource languages. In a low-resource settings, fine-tuning a pre-trained language model (PLM) is a feasible alternative to training a model from the scratch. However, it is a well-known fact that fine-tuning a model with a limited amount of data on a specific task can lead to overfitting. \citep{NEURIPS2021_e4a93f03} Therefore, we attempt to run several models, alternative to fine-tuning a PLM when evaluating our dataset, such as few-shot learning without prompts and traditional machine learning algorithms. We implement both deep learning and machine learning-based models with our dataset: fine-tuning a Korean PLM, Sentence Transformer Fine-tuning (SetFit; \citealp{tunstall2022efficient}), Support Vector Machines (SVM; \citealp{10.1023/A:1022627411411}), Logistic Regression (LR), Random Forest Classification (RFC), and Naive Bayes (NB). 

First, we fine-tune kobigbird-bert-base model\footnote{https://huggingface.co/monologg/kobigbird-bert-base} on our dataset. The dataset is randomly splitted into train, development, and test partitions at the 6:2:2 ratio. Fine-tuning is conducted on PyTorch 2.0.1 with Cuda 11.8, with a learning rate of 1e-5 and 5 training epochs. 

We also fine-tune a sentence-transformer model (paraphrase-mpnet-base-v2\footnote{https://huggingface.co/sentence-transformers/paraphrase-mpnet-base-v2}) with the SetFit approach. SetFit is a framework for few-shot learning fine-tuning of sentence-transformer models without using prompts. \citet{tunstall2022efficient} reports that this framework is more efficient and exhibits less variability when trained on a small dataset, compared to standard fine-tuning. SetFit utilizes two-stage training process. In the first process, sentence pairs are generated from the given data. These sentence pairs are used to fine-tune a sentence-transformer model. In the second process, the sentence pairs are encoded with the fine-tuned sentence-transformer model. These encodings are used to train the classification head. We follow these processes with paraphrase-mpnet-base-v2 model and our dataset. We randomly split our dataset into train and test sets at the ratio of 8:2. As SetFit only accepts a dataset consisting of two columns (text and label), we adjust the format of our dataset by concatenating the facts and claims. The following parameters are used to produce the final result: 32 batch size, 2 training epochs for body (first training process), 25 training epochs for head (second training process), and 3 iterations.

Lastly, we experiment with four different traditional machine learning algorithms: Support Vector Machines, Logistic Regression, Random Forest Classification, and Naive Bayes. When sufficient amount of data is given, deep learning-based models outperform traditional machine learning algorithms for many NLP tasks, such as classification. However, in a small data setting, applying traditional machine learning algorithms to the data can be a better approach than using deep learning-based models. Given the size of our dataset, we also implement four different machine learning models with our dataset.

As with the other models, the data labeled as 2 are removed. Then, preprocessing and tokenization are performed on the dataset using KoNLPy \cite{park2014konlpy}, in which process only content words, such as adjectives, verbs, and nouns, are saved. Thereafter, all the tokens are encoded as integers using CounterVectorizer. Then, the data are split to training and test data at the ratio of 8:2. The four machine learning-based models (i.e., Logistic Regression, Multinomial Naive Bayes, Support Vector Machine, and Random Forest Classifier) are implemented with a default setting and cross validation set to 10. Then, the models are optimized through grid search. 

We compare the results from the six different approaches to figure out which approach leads to the optimal result when there is no sufficient data. We also compare the results from our approaches to those of the Korean LJP benchmark model \cite{hwang2022a} to see if they are comparable.

\section{Results}
The evaluation result suggests that the models trained on our dataset can perform the task reasonably well. All the models/approaches, except for the Multinomial Naive Bayes, show accuracy scores higher than 63\%. Paraphrase-mpnet-base-v2 fine-tuned with SetFit outperforms all the other models (70.5\% accuracy), with the difference of six percent point or more in accuracy. The second best performing approach is Support Vector Machines, with 64.5\% accuracy. Kobigbird-bert-base model fine-tuned in a standard way places the third with the accuracy of 64.2\%. For the full evaluation results, see Table 2.

\section{Discussion}
The evaluation result confirms that SetFit demonstrates good performance even with small amount of data. It shows the best performance among the six models/approaches we experiment with. This result demonstrates that among SetFit, traditional machine learning-based approaches, and standard fine-tuning, SetFit is the best option when the amount of data is limited. The performance of traditional machine learning approaches and that of the standard fine-tuning approach do not show much difference. The result from Support Vector Machines is slightly better than the standard fine-tuning model, but its accuracy difference is marginal (0.3\%). All the other traditional machine learning approaches underperform the standard fine-tuning model, unlike our prediction.

The evaluation results also suggest that the models trained/fine-tuned on our dataset achieve performance comparable to the Korean LJP-Civil Lv1 benchmark models \citep{hwang2022a}, despite the small data size. The evaluation setting of the Korean LJP-Civil Lv1 benchmark models is very similar to our task. Its task is to predict the amount of money that the plaintiff would receive at the end of the trial. In the Lv1 task, the models are asked to predict the judgment with the three labels: 0, when the plaintiff is predicted to receive none, 2, when the plaintiff is expected to receive the whole amount as he/she claimed, and 1, when it is somewhere in between (0<x<1). It is designed as a multi-label classification task, but the class imbalance makes the setting similar to binary classification task (0: 2862, 1: 2132, 2: 87). Their dataset is significantly larger than our dataset (5081 cases vs. 473 cases), but the models trained on our dataset show the similar performance to the benchmark models trained on their dataset. The model fine-tuned on our dataset with SetFit approach shows 70.5\% accuracy, which is higher than the best benchmark model's score (69.1\% ± 0.1). This result demonstrates that it is possible to achieve a competitive performance with a small dataset, when sample efficient approaches (i.e., SetFit) are employed.

\section{Conclusion}
We present a Korean legal judgment prediction dataset, especially for legal disputes in the field of insurance. Recently, NLP techniques have been successfully applied to legal domains, and one such case is legal judgment prediction (LJP). There exist several LJP datasets, but these currently available datasets only concentrate on specific domains and large resource languages such as English or Chinese. We address this gap by introducing a Korean dataset in a novel and highly practical task. We also evaluate our dataset with various approaches, to figure out the alternative to standard fine-tuning in a small dataset setting. The evaluation result verifies that SetFit performs well in a small dataset setting. It is also observed that the models fine-tuned in SetFit approach on our dataset show similar performance to that of the Korean LJP benchmark models \citep{hwang2022a} despite a huge difference in the data size. This suggests that it is possible to achieve a competitive performance with a small dataset, when sample efficient approaches (i.e., SetFit) are employed.

\section*{Limitations}
The dataset presented in our study is rather limited in scope and size. It only contains Korean data from a very specific domain (i.e., insurance disputes). Given this limitation, it is possible that the findings from this study may be unique to this particular domain and task. They may not hold when extended to other languages, tasks, or fields. Findings from this study need further verification in more diverse settings. 

\section*{Ethics Statement}
All the insurance dispute cases contained in our datasets are fully anonymized, publicly available data. We made sure that there is no personal information in our dataset. Our dataset will be made publicly available with an open access. The dataset can potentially be used to predict the result of insurance disputes. However, the predicted results should not be considered as legal advice. When releasing our dataset, we will place a disclaimer stating that the user must not rely on any predictions made based on our dataset when making a legal decisions.

\bibliography{custom}
\bibliographystyle{acl_natbib}

\end{document}